\documentclass{article}

\usepackage{PRIMEarxiv}

\usepackage[utf8]{inputenc} 
\usepackage[T1]{fontenc}    
\usepackage{hyperref}       
\usepackage{url}            
\usepackage{booktabs}       
\usepackage{amsfonts}       
\usepackage{nicefrac}       
\usepackage{microtype}      
\usepackage{lipsum}
\usepackage{fancyhdr}       
\usepackage{graphicx}       
\graphicspath{{media/}}     

\usepackage{hyperref}
\usepackage{color}

\usepackage{graphicx}
\usepackage{caption}
\usepackage{subcaption}
\usepackage{xcolor}
\usepackage{colortbl}
\usepackage{courier}
\usepackage[leftmargin=3em]{quoting}
\usepackage{amsmath}
\usepackage{svg}

\usepackage{xargs} 
\usepackage[colorinlistoftodos,prependcaption,textsize=tiny, disable]{todonotes}     

\newcommandx{\unsure}[2][1=]{\todo[linecolor=red,backgroundcolor=red!25,bordercolor=red,#1]{#2}}
\newcommandx{\change}[2][1=]{\todo[linecolor=blue,backgroundcolor=blue!25,bordercolor=blue,#1]{#2}}
\newcommandx{\info}[2][1=]{\todo[linecolor=OliveGreen,backgroundcolor=OliveGreen!25,bordercolor=OliveGreen,#1]{#2}}
\newcommandx{\improvement}[2][1=]{\todo[linecolor=Plum,backgroundcolor=Plum!25,bordercolor=Plum,#1]{#2}}

\title{TinyML Towards Industry 4.0: Resource-Efficient Process Monitoring of a Milling Machine}

\author{
  Tim Langer\thanks{Work performed while at Fraunhofer Institute for Integrated Circuits IIS, Division Engineering of Adaptive Systems EAS, Dresden.}\\
  Chair of Highly-Parallel VLSI Systems\\and Neuro-Microelectronics\\
  TU Dresden\\
  \texttt{tim\_hauke.langer@tu-dresden.de}
  \And
  Matthias Widra\\
  Fraunhofer Institute for Machine Tools\\and Forming Technology IWU\\
  Dresden\\
  \texttt{matthias.widra@iwu.fraunhofer.de}
  \And
  Volkhard Beyer\\
  Fraunhofer Institute for Integrated Circuits IIS\\
  Division Engineering of Adaptive Systems EAS\\
  Dresden\\
  \texttt{volkhard.beyer@eas.iis.fraunhofer.de}
}

\begin{document}
\maketitle

\begin{abstract}
In the context of industry 4.0, long-serving industrial machines can be retrofitted with process monitoring capabilities for future use in a smart factory. One possible approach is the deployment of wireless monitoring systems, which can benefit substantially from the TinyML paradigm.
This work presents a complete TinyML flow from dataset generation, to machine learning model development, up to implementation and evaluation of a full preprocessing and classification pipeline on a microcontroller. After a short review on TinyML in industrial process monitoring, the creation of the novel MillingVibes dataset is described. The feasibility of a TinyML system for structure-integrated process quality monitoring could be shown by the development of an 8-bit-quantized convolutional neural network (CNN) model with 12.59\,kiB parameter storage. A test accuracy of 100.0\,\% could be reached at 15.4\,ms inference time and 1.462\,mJ per quantized CNN inference on an ARM Cortex M4F microcontroller, serving as a reference for future TinyML process monitoring solutions.
\end{abstract}


\section{Introduction}
Process monitoring can help to detect process faults early and thus reduce the rejection rates of manufactured products. However, in an established company, typically several large industrial machines have already been acquired long time ago. As the investment cost for an industrial machine tool is significant, there exists no urge to exchange the old machine with a more modern model with integrated process monitoring capabilities. As a cost-effective alternative, the machine can be modernized by retrofitting. Because retrofitting is sometimes only possible at space-limited or rotating locations within the machine, wired data and energy transmission might not be realizable. For an appropriate wireless implementation, a minimal energy consumption of the edge node is important. A minimal storage consumption of the machine learning (ML) model favors the selection of a low-power, low-cost microcontroller. Furthermore, it is desired to detect an erroneous process as fast as possible to avoid damage at the workpiece or even the machine tool. Therefore, a minimal latency until the classification is also considered a main goal. The required combination of low energy consumption, low storage consumption and low latency suggests the investigation of TinyML for structure-integrated sensing in the context on industry 4.0.

\section{Related Work}
To obtain an overview of research on resource-efficient machine learning for industrial process quality monitoring, the following keyword search string was used in Google Scholar:\\
\\
\textit{``tinyml'' AND (``condition monitoring'' OR ``industrial'' OR ``milling'' OR ``tooling machine'' OR ``anomaly'' OR ``anomaly detection'' OR ``process monitoring'' OR ``process quality monitoring''  OR ``production'' OR ``production process'' OR ``vibration'' OR ``accelerometer'')}\\
\\
The search lead to 843 results, of which 59 results were selected to be possibly relevant based on their headline (as of December 8th, 2022). The remaining articles were screened based on abstract and content, leading to 21 relevant articles, of which the most relevant approaches will be presented shortly.\\
\\

\subsection{TinyML for Different Industrial Applications}
\label{sec:tinyml_for_different_industrial_applications}
Among the relevant articles, three clusters focusing on different industrial use cases could be found, which will be described in the following.

\paragraph{Non-TinyML on Industrial Vibration Data}
Several approaches applied similar methodologies on industrial vibration data, but did not focus on resource efficiency following the paradigms of TinyML. Cinar et al. \cite{cinar_sensor_2022} proposed a solution using spectrogram-based classification of 3-axial vibration data for an electric motor testbed. Mey et al. \cite{mey_machine_2020} analyzed the unbalance of a rotating shaft with a convolutional neural network (CNN) on raw vibration data and with a fully-connected neural network on spectra.

\paragraph{TinyML for Anomaly Detection: } Antonini et al. \cite{antonini_tinyml_2022} investigated a vibration-based condition monitoring system especially suited for a retrofittable solution at a pump. However, due to harsh environmental conditions the data could not be recorded directly at the pump, but needed to be emulated using a computer fan testbed with injected vibrations. An interesting alternative to vibration-based anomaly detection has been presented by  \cite{abbasi_outliernets_2021} . The authors developed an auto-encoder architecture for acoustic anomaly detection of e.g. industrial fans and pumps, based on mel-spectrograms.

\paragraph{TinyML for Process Monitoring: } Two solutions were found investigating small-footprint ML for optical process monitoring \cite{bharti_edge-enabled_2022, pau_tiny_2021}. However, a major disadvantage of optical approaches is the high resolution required to see meaningful process variations, leading to large input data an complex models, sometimes prohibiting real-time execution or even implementation on a microcontroller with a high accuracy \cite{pau_tiny_2021}.

\begin{figure}
\centering
  \includegraphics[width=0.7\linewidth]
  {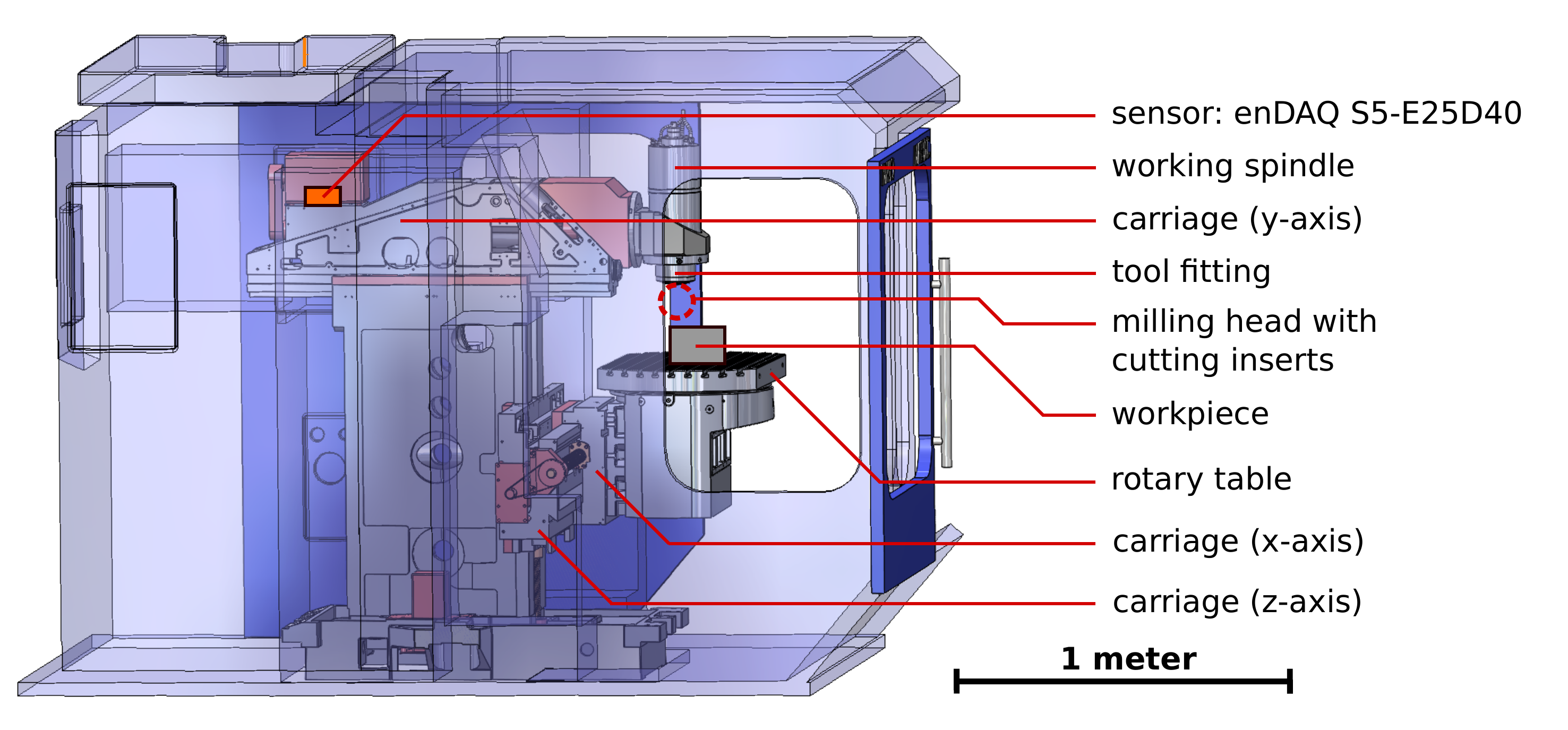}
  \caption{3D model of the Maho MH 800C machine tool for the experiments described in this article.}
  \label{fig:tooling_machine_model}
\end{figure}

\newpage

\subsection{Existing Datasets}
Several datasets for industrial processes on milling machines, machine tools, or ring bearings have already been published:

\begin{itemize}
    \item   Roughness of Milling Process Dataset \cite{yang_haw-ching_roughness_2020}
    \item   CWRU dataset \cite{case_western_reserve_university_cwru_2019}
    \item   NASA Milling Dataset \cite{agogino_best_2007}
    \item   NASA Bearing Dataset \cite{lee_ims_2007}
    \item   PRONOSTIA Dataset \cite{nectoux_pronostia_2012}
    \item   CNC Machining Data \cite{tnani_smart_2022}
    \item   MIMII Dataset \cite{purohit_mimii_2019}
\end{itemize}

\noindent However, these datasets generally have been recorded for condition monitoring of the current \textit{machine status}, not for a judgement of the \textit{process quality}. These datasets are mainly suitable for unsupervised anomaly detection, and thus not labeled for process quality detection of a material surface after milling. The novel dataset ``MillingVibes'' \cite{langer_millingvibes_2022} therefore delivers an important extension of publicly available datasets with focus on industrial process monitoring.

\section{MillingVibes: Dataset Generation for Milling Process Quality}

During the research summarized by this article, the MillingVibes dataset \cite{langer_millingvibes_2022} for a milling process with an industrial machine tool (see Fig. \ref{fig:tooling_machine_model}) has been created. The design of the experiment was guided by the question, whether an industrial machine tool can be retrofitted such that an early detection of a faulty milling process can be performed directly by the machine.
As the location for the retrofittable machine spare part has a large distance of approximately 1.40\,m from tool and workpiece, a relevant question was also whether process quality can be assessed at all from such a remote location.\\
To describe the creation of the MillingVibes dataset \cite{langer_millingvibes_2022}, the automated generation of milling patterns will be discussed first, followed by details on the dataset acquisition at the machine tool and the creation of a dataset from the vibration data of the milling process.

\subsection{Automated Generation of Milling Patterns}

\begin{figure}
\centering
    \centering
    \hspace*{\fill}
    \begin{subfigure}{0.3\linewidth}
        \centering
        \includegraphics[width=\linewidth]{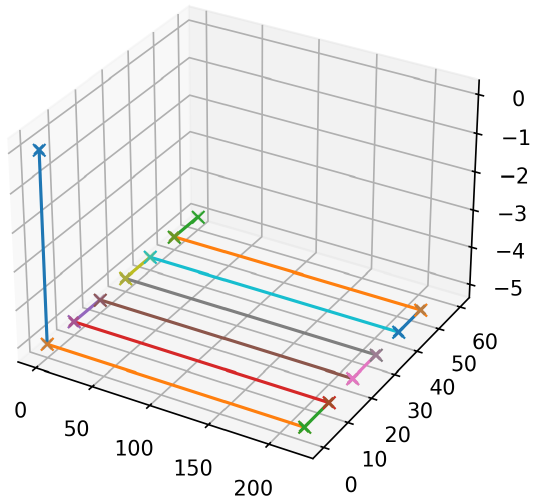}
        \caption{zig-zag pattern (virtual)} \label{fig:zig-zag_pattern__virtual}
        \label{fig:digital_verification_a}
    \end{subfigure}%
    \hspace*{\fill}
    \begin{subfigure}{0.38\linewidth}
        \centering
        \includegraphics[width=\linewidth]{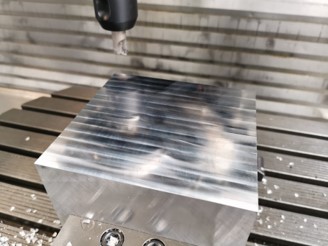}
        \caption{zig-zag pattern (actual)} \label{fig:zig-zag_pattern__actual}
        \label{fig:digital_verification_b}
    \end{subfigure}%
    \hspace*{\fill}
    \\
    \vspace{0.3cm}
    \hspace*{\fill}
    \begin{subfigure}{0.3\linewidth}
        \centering
        \includegraphics[width=\linewidth]{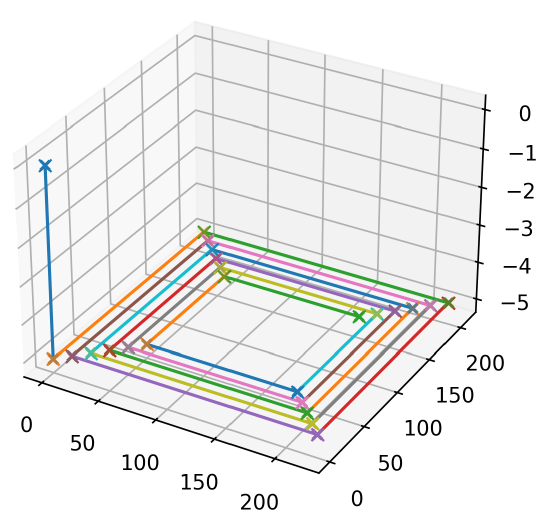}
        \caption{spiral pattern (virtual)} \label{fig:spiral_pattern__virtual}
        \label{fig:digital_verification_c}
    \end{subfigure}%
    \hspace*{\fill}
    \begin{subfigure}{0.38\linewidth}
        \centering
        \includegraphics[width=\linewidth]{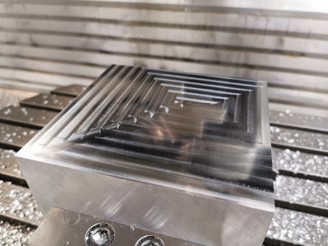}
        \caption{spiral pattern (actual)} \label{fig:spiral_pattern__actual}
        \label{fig:digital_verification_d}
    \end{subfigure}%
    \hspace*{\fill}
    \caption{Comparison of virtual and actual milling pattern.}
  \label{fig:digital_verification}
\end{figure}

To realize a variety of milling patterns, it was decided to automate the creation of G code scripts for the CNC console which controlled the milling machine. The automated G code generation was implemented in Python based on a combination of single linear movements, which could directly be translated to G code commands. A spiral pattern and a zig-zag pattern were chosen for the milling of a cuboid workpiece. While the spiral pattern offers the largest variety of axial movements (``\textpm x'', ``\textpm y''), the resulting axial movements were varying in length. On the contrary, one-axial movements (``\textpm x'') were recorded for the zig-zag pattern, which were constant in length (see Fig. \ref{fig:digital_verification_b}, \ref{fig:digital_verification_d}).\\
To verify the geometry of the milling pattern before possibly damaging the cost-intensive machine tool, a digital graphical simulation of the milling traces was performed a-priori using Python 3.10 and Matplotlib 3.6.2 (see Fig. \ref{fig:digital_verification}).

\subsection{Dataset Acquisition at the Tooling Machine}
\begin{figure}
\centering
\hspace*{\fill}
\begin{subfigure}{0.35\linewidth}
    \centering
    \includegraphics[width=\linewidth]{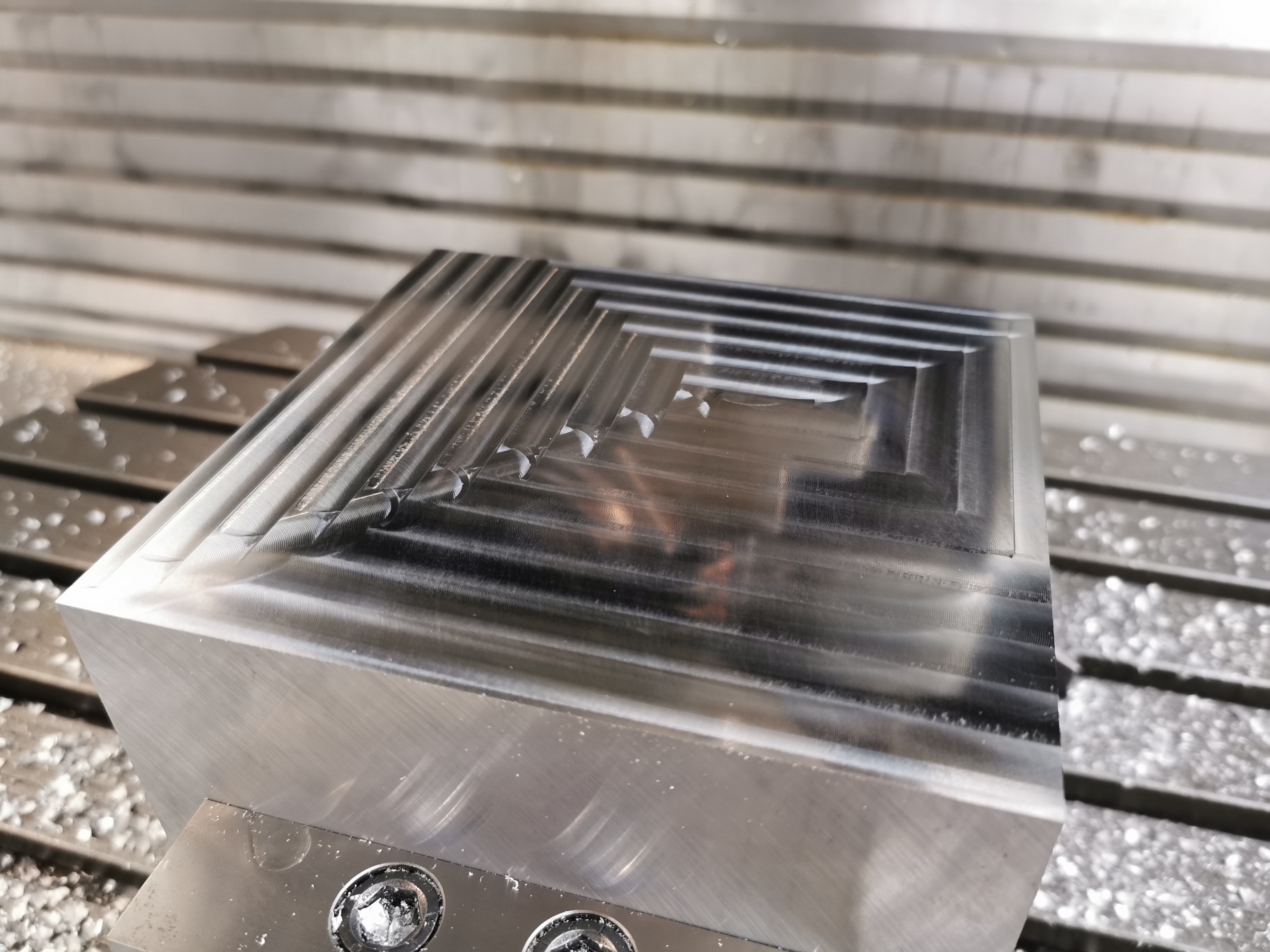}
    \caption{acceptable quality (smooth surface)} \label{fig:workpiece_good}
  \end{subfigure}%
  \hspace*{\fill}   
  \begin{subfigure}{0.35\linewidth}
    \centering
    \includegraphics[width=\linewidth]{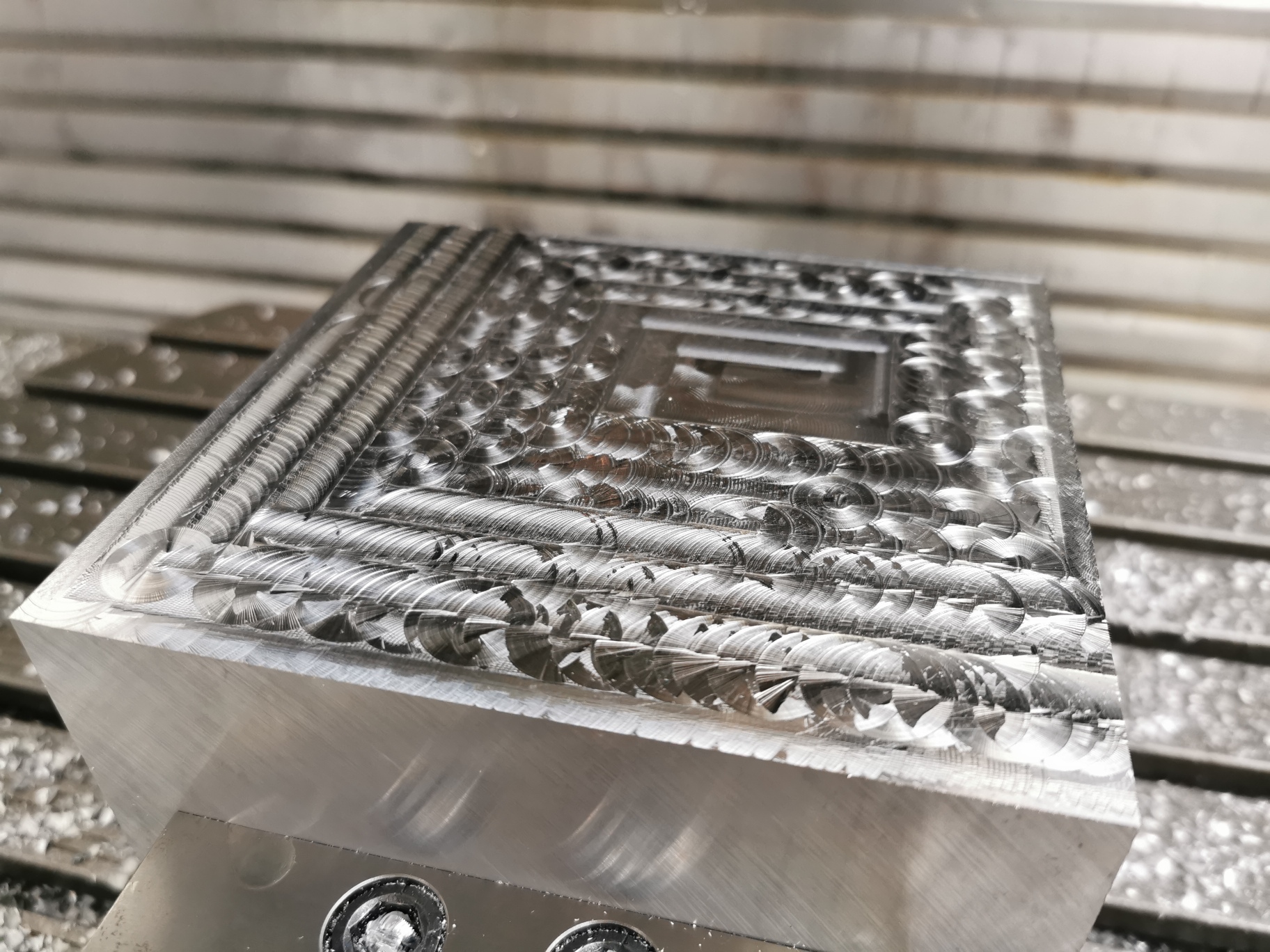}
    \caption{unacceptable quality (rough surface)} \label{fig:workpiece_bad}
  \end{subfigure}%
  \hspace*{\fill}
  \caption{Milling process quality of an aluminum workpiece.}
  \label{fig:workpiece_good+bad}
\end{figure}

The vibration data were recorded at a Maho MH 800C machine tool configured for a milling process (see Fig. \ref{fig:tooling_machine_model}). A Siemens WEISS 2SP1 working spindle was used, and the machine tool could be controlled via a Siemens Sinumerik 840D SL CNC control console with a control script using G code.\\
An aluminum alloy (EN AW 7075) was used as base material to simulate an industrial milling process with the machine tool as seen in Fig. \ref{fig:tooling_machine_model}. Each milling pattern was executed with a milling head of 20\,mm diameter, a constant rotational speed of 3000\,rpm, a milling depth of 2\,mm, and a constant forward feed. After a complete 2\,mm-thick aluminum layer was milled off the workpiece, 
further milling patterns were milled with constant forward feed rates from 960 to 3000\,mm/min.\\
Besides the generation of data samples with good surface quality after milling (see Fig. \ref{fig:workpiece_good}), different strategies were applied to create samples with bad surface quality (see Fig. \ref{fig:workpiece_bad}). The major strategy was to use an overly high forward feed, which lead to non-acceptable chatter marks. Additionally, the material was dry-milled (i.e.: without lubricant), which also favors the generation of surface defects. During experiments, the cutting inserts were also worn out and partly detached, which lead to a built-up edge and sometimes excessive chatter marks.\\
For data recording, a 3-axial, piezo-electric vibration sensor (enDAQ S5-E25D40) was positioned at the y-axis carriage at approximately 1.40\,m distance from the milling tool (see Fig. \ref{fig:tooling_machine_model}), and mechanically coupled to the machine with an adhesive tape. The three sensor and machine axes were aligned, and the vibration was sampled with 8,000\,Hz and 0.8\,mg resolution along all three axes.\\
In the MillingVibes dataset, a similar number of faulty and acceptable data samples could be realized, providing an advantage compared to other condition monitoring or anomaly detection datasets.
This was due to the investigation of a ``soft fault'' of surface roughness after processing, without serious consequences for the machine tool.



\subsection{Graphically Assisted Data Segmentation and Labeling}
\begin{figure}
\centering
    \includegraphics[width=0.6\linewidth]{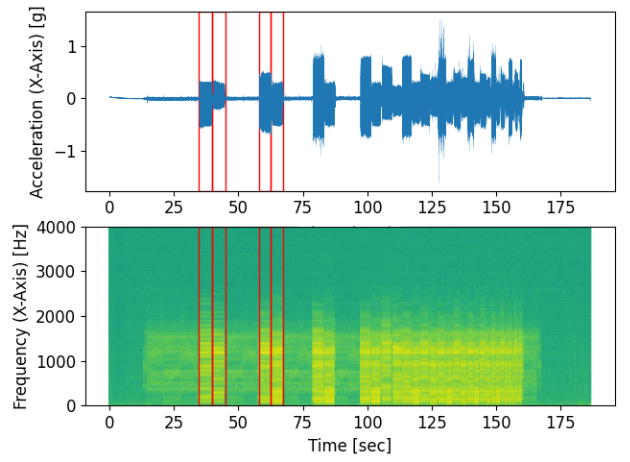}
    \caption{Signal segmentation and labeling for a spiral pattern in time and frequency domain.}
    \label{fig:signal_segmentation_time+freq_domain}
\end{figure}

The milling patterns were generated such that they covered more area than the surface of the workpiece, resulting in breaks without material contact between two linear movements for the zig-zag pattern. These breaks could be used to isolate the linear movement segments in both time and frequency domain. As continuous material contact was present for the spiral pattern, the precise border between two successive signal segments of the spiral pattern was harder to identify. A spectro-temporal display of time series data and the according spectrogram allowed a clearer separation of successive signal segments due to a spectral gap between two linear movements and the difference in spectral lines between the successive linear movements (see Fig. \ref{fig:signal_segmentation_time+freq_domain}). Based on the spectrogram, the signal segments were separated by manually creating a bounding box within Matplotlib, and the according time series signal segment was then exported with information on movement axis and direction along this axis (e.g. ``\textit{+x}''). Minor overlaps due to the graphical approach were later corrected, such that all linear movement segments were disjunct.

\subsection{Description of the Final Dataset}
After finishing the milling process, the surface quality of the workpiece was visually assessed as ``good'' or ``bad'' by a domain expert. The 3-axial time series data of the vibration sensor were segmented into isolated linear movements. Each signal segment was then divided into disjunct segments of 1\,sec length, to which the label ``good'' or ``bad'' was assigned depending on surface quality of the workpiece.  For example, a signal segment for a linear movement with material contact for 5.3\,sec was divided into 5 smaller segments of 1\,sec duration each (i.e.: 8000 samples at a sensor sampling rate of 8\,kHz). After isolation and labeling, the 1-sec-segments were randomly assigned to disjunct datasets for training and testing the TinyML model, with 872 \textit{(78.3\,\%)} segments used for training and 241 \textit{(21.7\%)} used for testing. Of the 872 training samples (test dataset: 241), 549 were labeled to have a good surface quality after the milling process (test dataset: 149), and 323 were labeled to have a bad quality (test dataset: 92).

\section{Development of a TinyML Classification Model}
An initial approach to use a CNN on the raw time-series vibration signal was not successful and was thus discarded early. Another time domain approach was based on the envelope of each vibration axis via Hilbert transform, as an initial hypothesis was that the amplitude was larger for a bad process quality due to more intense vibrations. However, the approach based on the envelope signal also did not lead to satisfying results. It was therefore decided to focus on frequency domain analysis based on the application of a CNN on spectrograms. Therefore, one spectrogram was calculated for each vibration axis, leading to a 3-dimensional input over time, frequency and vibration axis (see Fig. \ref{fig:cnn_architecture}).

\subsection{Preprocessing}
\label{sec:preprocessing}
To normalize the vibration data to the range [-1, 1], instance normalization was performed for each sample $\underline{\mathbf{x}}$ of dataset $\mathbf{X}$ as follows:
\begin{equation}
\underline{\mathbf{x}} = 2\, \cdot \, \frac{\underline{\mathbf{x}} - \textrm{min}\,\underline{\mathbf{x}}}{\textrm{max}\,\underline{\mathbf{x}} - \textrm{min}\,\underline{\mathbf{x}}} - 1 \;\; \forall \; \underline{\mathbf{x}} \in \mathbf{X}
\end{equation}
After normalization of the vibration data, a Tukey  window $\underline{\mathbf{w}}$ with window length $L_{Tukey} = N_{FFT} = $ 256 was applied to each input signal snippet to calculate a short-time Fourier transforms (STFT) frame, whereas two successive signal snippets for STFT were overlapping by 8 samples. Multiple STFTs were performed to generate a spectrogram for each vibration axis and each signal segment:
\begin{equation}
X(n, k)\, =\sum_{m\,=\,-\infty}^{+\infty}w[n-m]\,\cdot\,x[m]\,\cdot\,\exp\left(\frac{ j\cdot2\pi \cdot k}{N_{FFT}} \cdot m\right)
\end{equation}
with $n, k$ as indices for temporal and spectral position within the spectrogram, $m$ as index for its calculation within the time-domain signal $\underline{\mathbf{x}}$ and Tukey window $\underline{\mathbf{w}}$, and $N_{FFT}$ being the Fast-Fourier Transform (FFT) order for calculating the spectrogram. The resulting spectrogram of each 1-second signal dataset sample has 32 timesteps with 129 frequency bins each for all 3 vibration axes.\\
After reducing the feature map size with a (8x2)-average-pooling step, the spectrogram was log-transformed to emphasize the magnitude differences between spectral lines compared to background noise before CNN-based classification:
\begin{equation}
X_{log}(n, k)\, = \textrm{log\textsubscript{10}}\; X(n, k)
\end{equation}
While the previous steps were performed in the floating-point domain to ensure numerical precision, the feature map was then converted to an INT8 representation to ensure computationally efficient CNN inference.

\subsection{Development of a Machine Learning Model}
After discarding the usage of a CNN on the raw vibration signal or its envelope, it was decided to train CNNs on data preprocessed according to section \ref{sec:preprocessing}. Therefore, a grid search was performed to find both architecture and hyperparameters of the best-performing CNN. The CNN models were trained using Keras with batch size 16 for maximum 200 epochs with an early stopping strategy (patience: 5 epochs). An Adam optimizer was used with an initial learning rate of $5\textrm{x}10^{-4}$ and exponential learning rate scheduling (decay rate: 0.95 every 50 training steps). Before training the model and shuffling the training dataset, all relevant Python packages were seeded to ensure reproducibility.\\
As an advantage over other datasets for predictive machine maintenance or process monitoring, the MillingVibes dataset \cite{langer_millingvibes_2022} contains a relatively balanced number of good and bad samples, since the bad samples could be generated without severe implications towards the milling machine. Therefore, accuracy could be taken as metric to judge CNN model performance, opposed to extremely imbalanced datasets.\\
The CNN model as seen in Fig. \ref{fig:cnn_architecture} achieved the highest validation accuracy among the investigated models, and could reach 100.0\,\% on the test dataset, which is likely due to the relative simplicity of the presented task.


\subsection{Model Adaptation for Deployment on a Microcontroller}
To minimize the required storage footprint and inference time, it was decided to quantize all model weights to signed 8-bit integer values (INT8) using the post-training quantization of TFLite. The intermediate activations were also quantized to INT8 using the TFLite flow based on datapoints randomly sampled from the training set. Additionally, the input layer of the CNN was quantized to INT8. The fully-quantized CNN model was then exported as a *.tflite file and could maintain the achieved floating-point test accuracy of 100.0\% on the desktop PC. After testing, the TFLite model was converted for an ARM Cortex M4F microcontroller using the STM32Cube.AI tool within STM32Cube IDE. Eventually, the C code of the CNN model could be generated automatically and the model weights could be exported into a C header file, using the STM32Cube.AI runtime for inference on the microcontroller.


\section{Performance of the Embedded Implementation}
To investigate the suitability of a retrofittable TinyML edge node for industrial process quality monitoring, both preprocessing and CNN model were implemented on an ARM Cortex M4F microcontroller on an STM32L4R5ZI Nucleo prototyping board. The processor was configured to run at a clock frequency of $f_{CLK} =$ 120\,MHz. To implement the CNN model on the microcontroller, STM Cube IDE v1.10.0 was used in combination with the machine learning environment Cube AI v7.3.0 and GCC v11.3.0 with optimization level \texttt{-Ofast} for C code compilation.

\begin{figure*}
\centering
  \includegraphics[width=\linewidth]{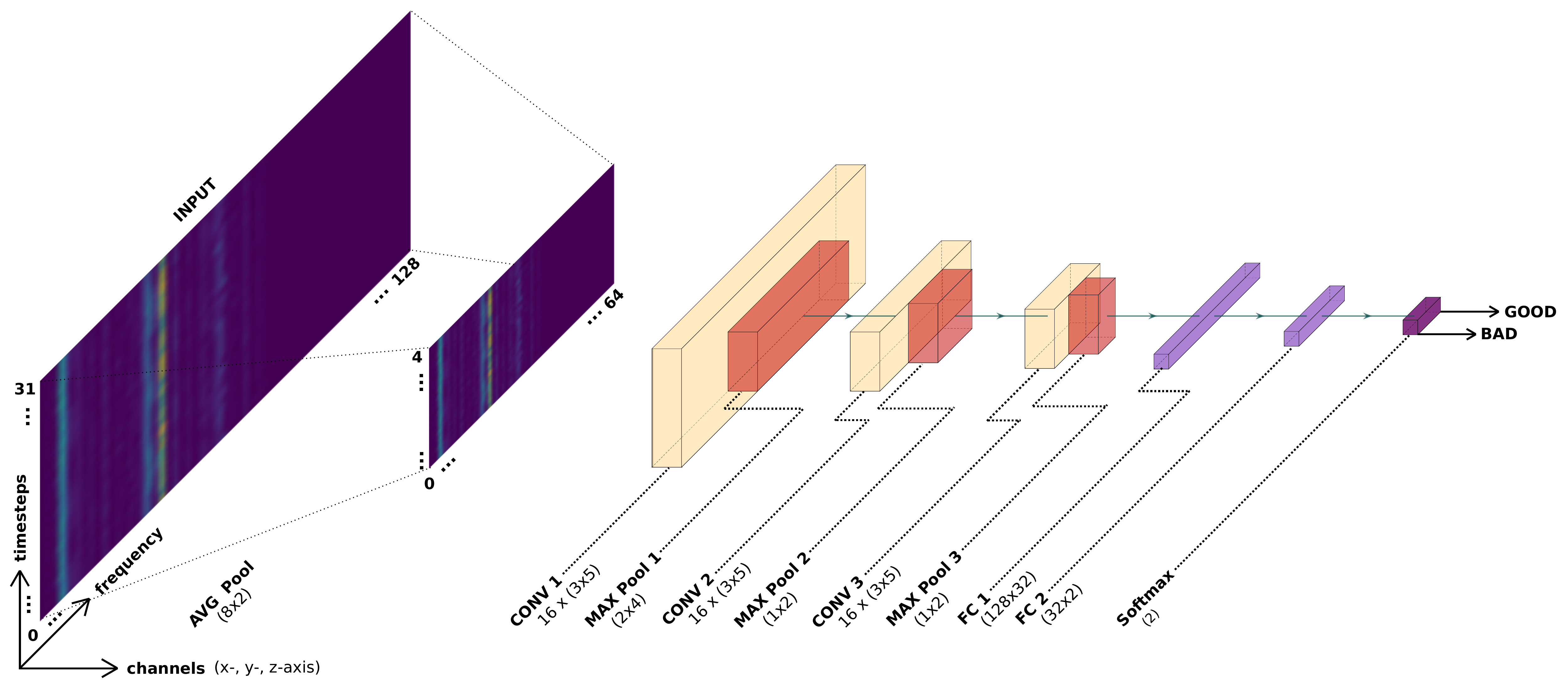}
  \caption{CNN architecture for classification of 3-axial accelerometer spectrogram segments (original signal duration: 1\,sec, graph generated with PlotNeuralNet \cite{iqbal_harisiqbal88plotneuralnet_2018}).}
  \label{fig:cnn_architecture}
\end{figure*}

%

\subsection{Memory Usage}
To reduce the memory footprint as well as inference time, the CNN model was optimized to have extremely few parameters, following the TinyML paradigms. The CNN architecture as seen in Fig. \ref{fig:cnn_architecture} requires 12.59\,kiB flash memory for parameter storage, and additionally 39.75\,kiB for the provided runtime library, resulting in 51.34\,kiB CNN-related flash usage. Activations and library require 7.72\,kiB and 4.37\,kiB RAM storage, respectively, summing up to 12.09\,kiB CNN-related RAM usage.\\
Additional memory usage results especially from the input signal (3 axes of 8,000 32-bit floating-point samples each), requiring 93.75\,kiB flash and RAM memory, respectively, since the sample data has been provided as header file. Also, look-up tables and coefficient arrays for FFT consume a noticeable amount of flash memory.\\
In total, 112.58\,kiB of 640\,kiB available RAM, and 253.37\,kiB of 2\,MiB available flash memory are required.

\subsection{Preprocessing and Inference Runtime}
The runtime of different preprocessing and inference subtasks (see Tab. \ref{tab:runtime_per_subtask}) was measured using an internal timer of the ARM Cortex M4F microcontroller. Multiple iterative optimization steps for preprocessing and model re-design were performed to achieve low runtimes of the individual subtasks, while maintaining a high classification accuracy at a low storage budget. By these adaptations, the CNN inference time could be reduced from 845\,ms to 15.4\,ms, and the preprocessing runtime could be reduced from 467\,ms to 69.1\,ms. The major time is consumed by the spectrogram calculation, logarithmic (LOG) scaling, and the CNN inference. To reduce the impact of LOG-scaling, it was decided to introduce average (AVG) pooling as a preprocessing step after computing the spectrogram. By this, the input feature map size was reduced by a factor of 16 from \textit{3x32x129} to \textit{3x4x65}.\\
Additional time savings could be achieved by using ARM-CMSIS vector instructions. As an example, min-max-scaling for approximately 24,000 values took 7.1\,ms using CMSIS-DSP instructions compared to 8.7\,ms for the hand-crafted implementation, which is equivalent to a 18.3\% reduction in execution time.

\subsection{Energy Consumption}
The power consumption of the TinyML model was calculated based on the voltage difference $V_1[t] - V_2[t]$ over a shunt resistor $R_{shunt}$, and the supply voltage $V_2[t]$ of the microcontroller. A relatively large shunt resistor of $R_{shunt}=10\,\Omega$  was chosen to allow a good resolution for voltage measurements due to the small current $I_{DD}$. The shunt resistor was connected to jumper JP5 on the STM32 Nucleo L4R5ZI prototyping board, which is intended for $I_{DD}$ measurements.  The voltages $V_1[t]$, $V_2[t]$ were sampled at 16-bit resolution with a frequency of $f = 10\,\textrm{kHz}$ using a MEphisto Scope UM203, while continuously running the CNN preprocessing and inference for $T=10\,\textrm{sec}$. Supply voltage $V_{DD}[t]$, supply current $I_{DD}[t]$, average power $P_{avg}$ and energy per inference (EPI) $EPI_{avg}$ could then be derived as follows:
\begin{align}
    V_{DD}[t] &= V_2[t] \label{eq:v_dd} \\
    I_{DD}[t] &= \frac{V_1[t] - V_2[t]}{R_{shunt}} \label{eq:i_dd} \\
    P_{avg}   &=  V_{DD, avg} \cdot I_{DD, avg} \label{eq:p_avg}\\
    EPI_{AVG} &= P_{avg} \cdot t_{pre+CNN,avg} \label{eq:epi_avg}
\end{align}

\noindent with $t_{pre+CNN,avg}=84.5$\,ms as average time for a single CNN inference including preprocessing. At a core clock frequency of $f_{CLK}=$ 120\,MHz with average voltage $V_{DD,avg}\approx$ 2.987\,V
 and average current $I_{DD,avg}\approx$ 31.807\,mA, an average power consumption of $P_{avg} \approx$ 94.9\,mW was calculated for the ARM Cortex M4F core of the Nucleo L4R5ZI prototyping board. Following Eq. \ref{eq:epi_avg}, the average EPI could be estimated as $EPI_{avg} \approx$ 8.022\,mJ. Considering only the CNN inference without preprocessing, the model achieved $EPI_{avg} \approx$ 1.462\,mJ at an average inference time of 15.4\,ms.

\begin{table}
\centering
  \caption{Preprocessing and Inference Runtime}
  \label{tab:runtime_per_subtask}
  \begin{tabular}{lr}
    \toprule
    \multicolumn{1}{l}{\textbf{Subtask}} & \multicolumn{1}{c}{\textbf{Runtime}}\\ \hline
    \rowcolor{gray!25}
    input normalization to [-1, +1]         &   7.2\,ms                  \\ 
    spectrogram calculation (3 axes)        &   30.6\,ms                 \\
    \rowcolor{gray!25}
    input compression (AVG pooling)          &   3.9\,ms             \\
    LOG-scaling spectrogram                 &   26.9\,ms                \\
    \rowcolor{gray!25}
    transposition for CNN                   &   0.2\,ms                \\
    INT8 quantization to [-128, +127]       &   0.3\,ms    \\
    \rowcolor{gray!25}
    CNN inference                           &   15.4\,ms           \\
\end{tabular}
\end{table}

\section{Discussion}
The presented MillingVibes dataset \cite{langer_millingvibes_2022} complements existing industrial datasets, as it focuses on the process quality, provides labeled data, and has a relatively balanced amount of good and bad samples. The acquired dataset served as the foundation for the development and in-depth benchmarking of a TinyML model including preprocessing, which were both deployed on a commercial microcontroller. Compared to other TinyML models like presented by Cinar et al. \cite{cinar_sensor_2022}, Bharti et al. \cite{bharti_edge-enabled_2022} or Pau et al. \cite{pau_tiny_2021}, the presented solution has a low amount of parameters with only 12.59\,kiB required parameter storage. Especially compared to similar approaches that do not explicitly focus on resource-efficient computing like presented by Mey et al. \cite{mey_machine_2020}, the relation of invested memory and obtained accuracy can be highlighted. Although the model is small with 12.59\,kiB required parameter storage, the STM32Cube.AI runtime library requires more than 3x the parameter storage, for which the implementation as bare metal C code is recommended for a minimal storage footprint in the future.\\
While a binary classification problem has a low complexity in general, the achieved test accuracy of 100.0\% of the 8-bit-quantized model is remarkable and could only be achieved after several iterations of optimization. It was found that switching from dataset normalization (min/max of all data samples) to instance normalization (min/max per sample) was especially beneficial for performance of the quantized and non-quantized models. One conclusion of this is that the information provided by magnitude relations between different dataset samples can be ignored, since it is equalized by instance normalization. Instead, the location of the spectral lines and the magnitude relations within a single sample (as emphasized by instance normalization) provide highly relevant information.\\
Often, TinyML implementations focus on the performance of the machine learning model alone. In contrast, this publication holistically also considers the embedded implementation of the preprocessing steps, breaking down the required time for each preprocessing step. Initially it was assumed that the forward pass of the CNN model would require the major proportion of the complete inference (including preprocessing). However, it was found that especially heavily-repetitive and complex operations like the spectrogram generation or just computationally very inefficient operations like logarithm computations consume the major shares of the complete inference time budget, especially with CNNs heavily optimized for embedded inference. This emphasizes the necessity for thorough consideration of preprocessing steps, and their co-optimization with the machine learning model. With an overall inference time of approximately 84.5\,ms, the presented approach allows a fast response compared to the typical time constants for milling processes.\\
While there are microcontroller-based TinyML solutions existing with a lower EPI in other domains, often the energy consumption is not even measured, especially for TinyML applications related to process monitoring. This article therefore also gives a good orientation for other works, since it focuses on the three most significant optimization criteria of TinyML (besides classification accuracy): storage, latency and energy consumption. While each of the three criteria can possibly be improved (depending on the optimization goal), this work presents a good trade-off for all categories. The special value of this publication lies in the end-to-end approach from dataset generation to performance measurement on the microcontroller, and can thus also serve as a reference for different domains.


\section{Conclusion}
In this article, we presented an end-to-end case study for the implementation of a TinyML model for industrial process monitoring, using a milling process as example. The presented end-to-end flow covers dataset acquisition, development and optimization of a CNN including preprocessing for an embedded system, as well as performance benchmarking on a commercial low-power microcontroller. It could be shown that even with a sensor attached to a remote machine part, process quality issues can be detected early with an edge node based on a microcontroller. The INT8-quantized CNN model could maintain a test accuracy of 100.0\% and perform a single inference in 15.4\,ms (84.5\,ms with preprocessing) using only 1.46\,mJ (8.02\,mJ) of energy on a commercial ARM Cortex M4F microcontroller, requiring only 12.59\,kiB plain parameter storage and 7.72\,kiB for intermediate activations.\\
This exemplary use case shows the suitability of TinyML systems for industry 4.0 edge nodes and the potential of retrofitting existing machine tools with additional smart sensor technology. Possible future research directions include the extension of the presented task from a binary process quality decision (``good'' vs. ``bad'') to more complex analyses. Examples of advanced analyses are the determination of the process quality fault type, detection of machine tool configuration (such as forward feed or rotation speed) or detecting critical movement patterns for process quality. Additionally, further optimizations (bare metal C code instead of STM32Cube.AI runtime, using low-power FPGA accelerators, etc.) could be investigated to achieve minimal storage, latency and energy requirements. Another path to even more efficient algorithm implementations can be to investigate the co-optimization of neural classifiers and preprocessing algorithms by hardware-aware neural architecture search.

\section*{Acknowledgements}
The authors would like to thank Frank Arnold at Technische Universität Dresden, Chair of Machine Tools Development and Adaptive Controls, for his input on possible process variations to generate varying surface quality during milling. This work was financed by the Saxon Federal Government out of the state budget approved by the Saxon State Parliament and the Federal Ministry of Education and Research of Germany.

\bibliographystyle{IEEEtran}
\bibliography{IEEEabrv, MAIN}

\end{document}